# An NKCS Model of Bookchin's Communalism

Larry Bull

Department of Computer Science & Creative Technologies

University of the West of England, Bristol UK

Larry.Bull@uwe.ac.uk

**Abstract**

The NKCS model was introduced to explore coevolutionary systems, that is, systems in which multiple species are closely interconnected. The fitness landscapes of the species are coupled to a controllable amount, where the underlying properties of the individual landscapes are also controllable. No previous work has explored the use of hierarchical control within the model. This paper explores the effects of using a confederation, based on Bookchin's communalism, and a single point of global control. Significant changes in behaviour from the traditional model are seen across the parameter space.

**Introduction**

Kauffman and Johnsen [1992] presented a tuneable, abstract model through which to explore some of the basic properties of coevolutionary systems – the NKCS model. Extending the well-known NK model [Kauffman and Levin, 1987] of rugged fitness landscapes, the NKCS model couples multiple NK landscapes to study the evolutionary dynamics of ecosystems containing multiple species. Versions of the model have also been applied to exploring complex systems, such as in management science (e.g., [Levitan et al., 1997]).

Kauffman [1995, pp245-271] later presented a version of the model in which the ecosystem is placed upon a two-dimensional grid. At one extreme, each cell – termed a patch - represents a separate species coevolving with its eight immediate neighbours, and at the other, the whole grid is one species. Shifting to social systems as the subject of study, these are described as abstract representations of extreme capitalism and communism respectively. It is shown how, as patch size is increased from small to large, the fitness of the overall system increases and then decreases; a critical degree of autonomy appears to exist.

Although not explicitly stated, a version of the NK model has been presented which enables varying degrees of the wider consideration of the effects of local action/change in decentralised systems that was loosely inspired by social anarchism – termed the NKD model [Bull, 2020]. The degree and constituency of the connected/social considerations were shown as critical, potentially more efficient than centralised/global control.

Drawing upon both anarchism and communism, Bookchin's [2015] communalism proposes a social structure consisting of autonomous assemblies working within a confederated structure wherein majority voting at the representation level is the mechanism through which local action/change is considered more widely. This structure is amenable to simple modelling using the NKCS model with assemblies represented by the species and an overriding confederation layer using majority voting determining the acceptance (or not) of changes within species. That is, the effects of adding hierarchy to the NKCS model can be explored and the resulting behaviour compared with the original. This paper shows how Bookchin's communalism proves beneficial over coevolving assemblies for larger, more interdependent systems.

**The NKCS Model**

Kauffman and Levin [1987] introduced the NK model to allow the systematic study of various aspects of fitness landscapes. In the standard model, the features of the fitness landscapes are specified by two parameters: $N$, the length of the genome; and $K$, the number of genes that has an effect on the fitness contribution of each (binary) gene. Thus, increasing $K$ with respect to $N$ increases the epistatic linkage, increasing the ruggedness of the fitness landscape. The increase in epistasis increases the number of optima, increases the steepness of their sides, and decreases their correlation [Kauffman, 1993]. The model assumes all intragenome interactions are so complex that it is only appropriate to assign random values to their effects on fitness. Therefore for each of the possible $K$ interactions a table of $2^{(K+1)}$ fitnesses is created for each gene with all entries in the range 0.0 to 1.0, such that there is one fitness for each combination of traits. The fitness contribution of each gene is found from its table. These fitnesses are then summed and normalized by $N$ to give the selective fitness of the total genome.

Kauffman and Johnsen [1992] subsequently introduced the abstract NKCS model to enable the study of various aspects of *co*evolution. Each gene is said to also depend upon $C$ randomly chosen traits in each of the other $S$ species with which it interacts. Altering $C$, with respect to N, changes how dramatically adaptive moves by each species deform the landscape(s) of its partner(s), where increasing C typically increases the time to equilibrium. Again, for each of the possible $K+(S \times C)$ interactions, a table of $2^{(K+(S \times C)+1)}$ fitnesses is created for each gene, with all entries in the range 0.0 to 1.0, such that there is one fitness for each combination of traits. Such tables are created for each species (Figure 1).

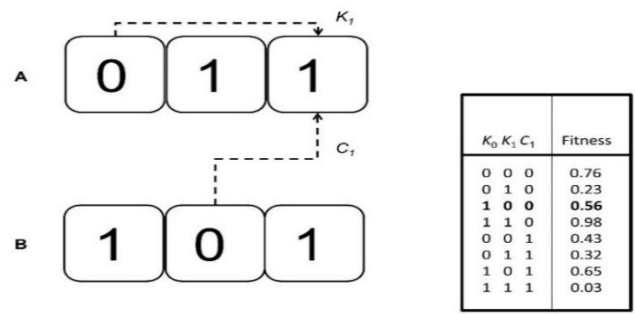

Fig. 1. The NKCS model: Each gene is connected to K randomly chosen local genes and to C randomly chosen genes in each of the S other species. Connections and table shown for one gene in one species for clarity. Here N=3, K=1, C=1, S=1.

Following [Kauffman, 1993], a random hill-climbing algorithm is used here to examine the properties and evolutionary dynamics of NKCS models. That is, a species evolves by making a random change to a randomly chosen gene per generation. The "population" is said to move to the genetic configuration of the mutated individual if its fitness is greater than or equal to the fitness of the current individual.

Figure 2 shows example behaviour for one of two coevolving species where the parameters of each are the same and hence behaviour is symmetrical. The effects of mutual fitness landscape movement are clearly seen.

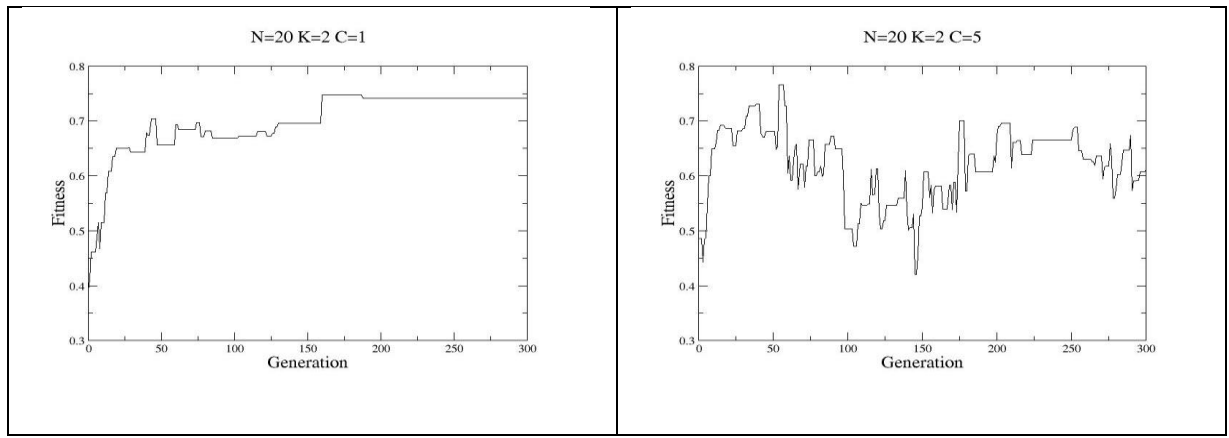

Fig. 2. Showing example single runs of the typical behaviour of the standard NKCS model of coevolution with different degrees of landscape coupling (*C*), with two species (*S*=1).

**Confederated Assemblies in the NKCS Model**

As noted above, the species of the NKCS model can be mapped directly to assemblies consisting of *N* components (individuals), each depending upon each other to some degree (*K*) and upon others within the overall social system (*S*x*C*). A second layer is added where proposed changes are voted upon. On each iteration, the next assembly in turn makes a random alteration and if that maintains or increases the fitness of the given assembly, the effect on the fitnesses of the other *S* assemblies is calculated. If the fitnesses of the majority of the other assemblies is maintained or improved, the change is adopted. For example, with *S*=2, if one or both of the other assemblies essentially vote in favour of the change (since there is at least no detrimental effect upon them), the change is adopted. A central control variant is also explored wherein a change must be at least neutral for all assemblies.

All results reported in this paper are the average of 10 runs (random start points) on each of 10 randomly created NKCS fitness landscapes, i.e., 100 runs, for 20,000 generations, for each parameter configuration. The (average) final fitness of the whole system is used for comparisons. Here 0≤*K*≤8, 1≤*C*≤3, *S*=2, for *N*=20 and *N*=100. The total fitness of the three species is divided by *S*+1 to maintain the traditional fitness range [0.0,1.0].

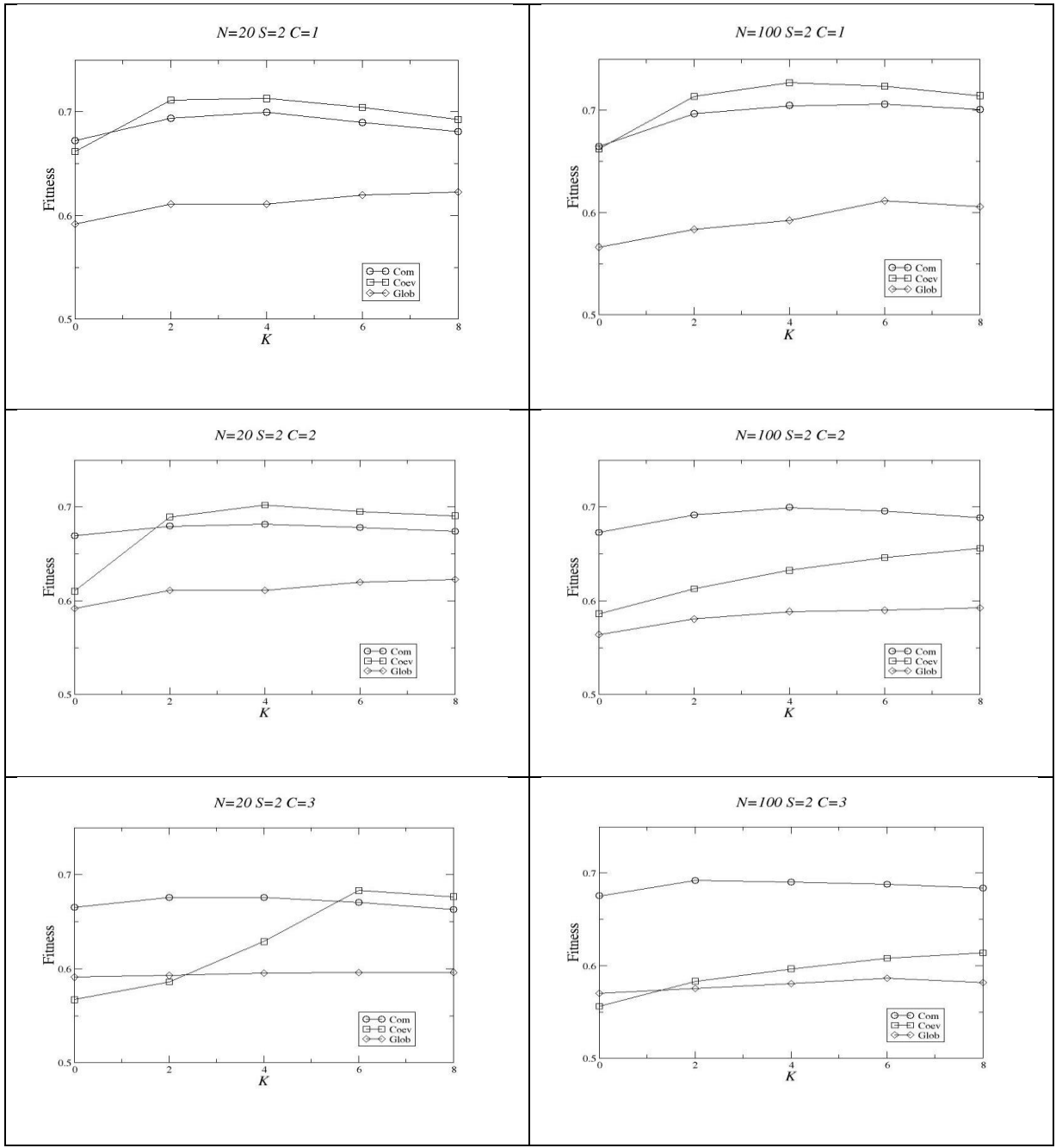

Fig. 3. Showing the total fitness reached by all assemblies after 20,000 generations on landscapes of varying ruggedness (*K*), coupling (*C*), and length (*N*). Communalism (Com), standard coevolution (Coev), and a central controller (Glob) are compared.

Figure 3 shows how conditions exist under which all three social structures prove beneficial and that communalism is beneficial for larger $N$ with higher values of C in particular. For small $N$ and $C$=1, coevolution is more efficient than communalism for $K>0$ (T-test, $p<0.05$), and both are significantly better than global/centralised control under all K (T-test, $p<0.05$). For higher $C$=3, communalism is the most efficient system when $K<6$ (T-test, $p<0.05$), with coevolution the best for higher $K$ (T-test, $p<0.05$). That is, coevolution performs better than communalism when $K>(SxC)$ here. This critical regime for coevolutionary systems was identified by Kauffman and Johnsen [1992]. Global control is more efficient than coevolution when $K<2$ (T-test, $p<0.05$), the same for $1<K<4$ (T-test, $p\geq0.05$) and significantly worse thereafter, and always worse than communalism (T-test, $p<0.05$).

For larger $N$ and $C>1$, communalism is significantly more efficient than coevolution and global control for all $K$ tried (T-test, $p<0.05$). That is, in particular, the critical regime for coevolution identified when $N$=20 disappears; larger coevolutionary systems appear more sensitive to interdependence (Kauffman and Johnsen [1992] used $N$=24). The relative performance of global control and coevolution is generally the same as for $N$=20 when $C$=3.

Thus, the additional hierarchical layer of a majority vote confederation in the NKCS model proves beneficial over a significant proportion of the parameter space. Unbridled coevolution remains beneficial over a larger proportion of the parameter space however. As Alexander Pope wrote, "To err is Humane" and the model above has assumed that individual assemblies are able to make accurate estimates of the effects of the proposed change by the others upon themselves.

Figure 4 shows the effects of adding a degree of error to that decision process. That is, in the case when all $S$ unchanging assemblies calculate the change would be detrimental to themselves, they still vote in its favour a percentage of the time. Clearly, this was 0% of the time above. As can be seen, communalism improves in its efficiency under almost all conditions where coevolution was previously the most beneficial scenario when 10% or 25% of decisions are erroneous. However, this comes at the cost of reduced efficiency where it was previously the better scheme with error rates used here. It may be that can be mitigated somewhat by fine-tuning the error rate but it has not been explored further here.

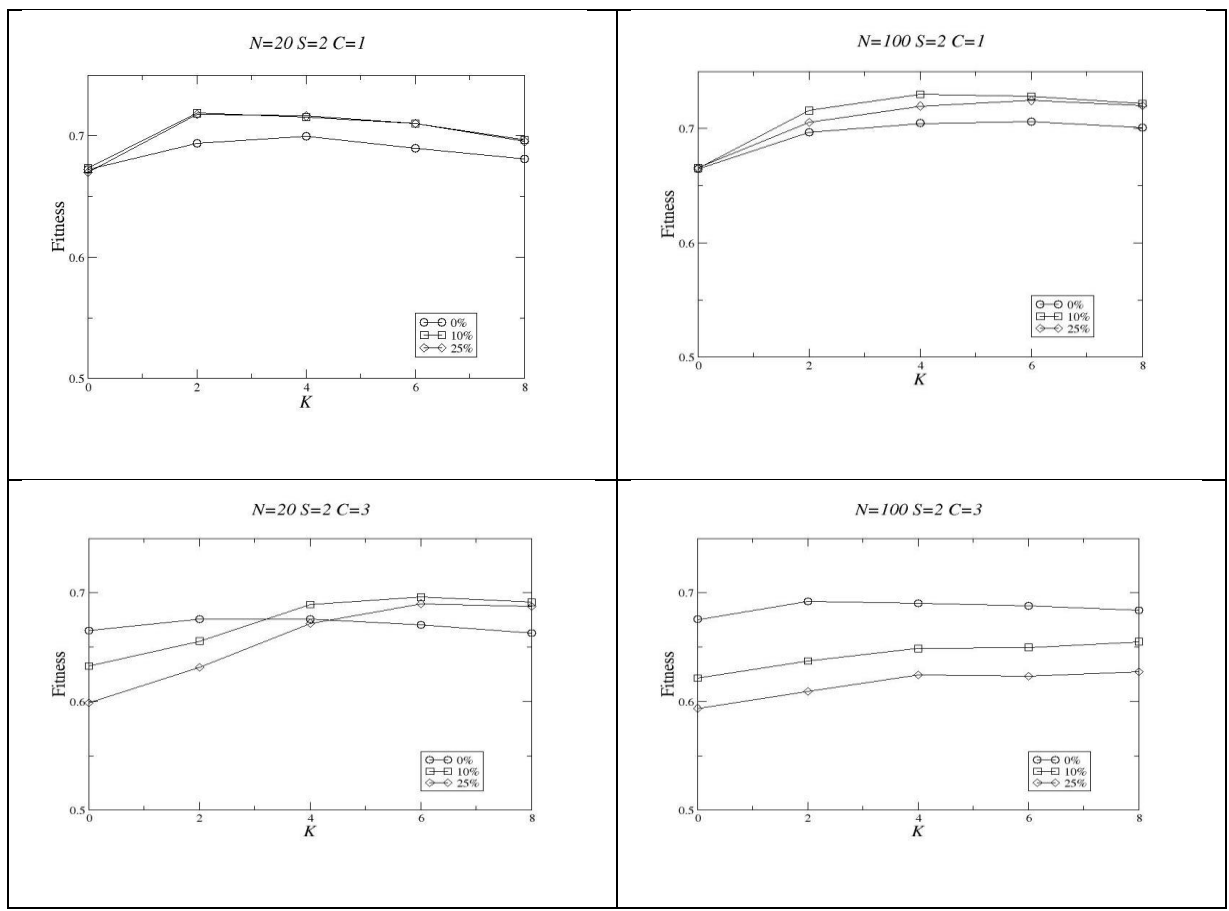

Fig. 4. Showing the total fitness reached by all assemblies after 20,000 generations under communalism with varying degrees of error in the estimations of effects of assembly changes.

Finally, Bookchin [2015] notes the possibility – if not inevitability – of assemblies being of different sizes due to the most natural/logical way in which they would be formed within local areas. It has previously been shown how using asymmetrically sized species can affect behaviour within the NKCS model [Bull, 2021]. Figure 5 shows example behaviour of varying the relative size of assemblies whilst maintaining the same overall number of entities/people as for N=100 above. The example of two small and one large assembly is shown, as is two large and one small. As can be seen, communalism also appears sensitive to the relative size of its constituents, with equally sized assemblies appearing to generally be the most robust.

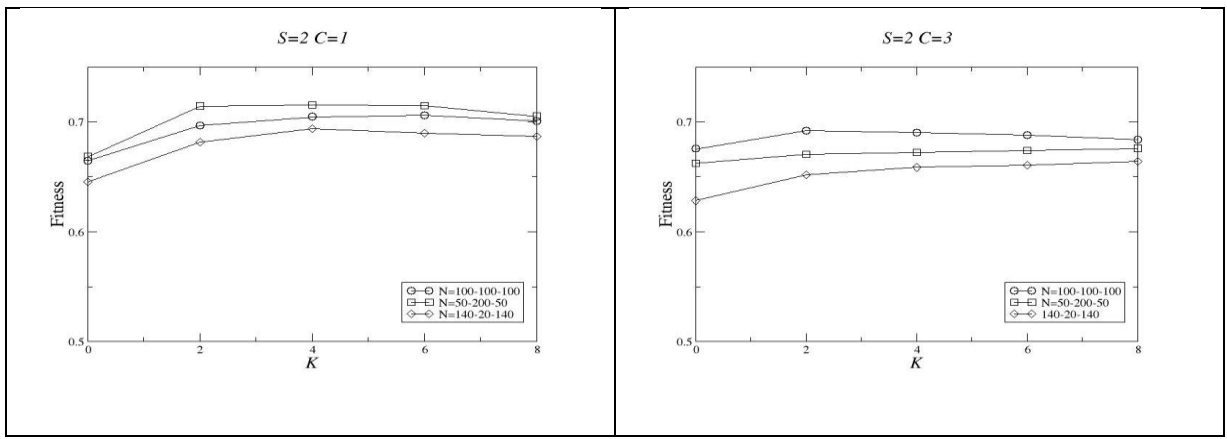

Fig. 5. Showing the total fitness reached by all assemblies after 20,000 generations under communalism with differently sized assemblies when $S=2$.

**Conclusion**

The original NKCS model - and all known uses of it thereafter – does not include hierarchical control. It has been shown here that introducing a confederation can be beneficial, depending upon the degree of landscape ruggedness ($K$), the degree of landscape coupling ($C$), and the size of the partners' genomes ($N$) in comparison to the traditional case.

Future work will further consider confederation mechanisms beyond simple majority voting, based on the added decision error findings reported here, as well as the use of hierarchies of confederations to potentially improve scalability.